\documentclass[10pt,twocolumn,letterpaper]{article}

\usepackage[pagenumbers]{cvpr} 

\usepackage{indentfirst}

\def\1{mathbb{1}}

\def\I{{\bf I}}

\def\0{{\bf 0}}
\def\1{{\bf 1}}

\makeatletter
\DeclareRobustCommand\onedot{\futurelet\@let@token\@onedot}
\def\@onedot{\ifx\@let@token.\else.\null\fi\xspace}

\definecolor{purple}{rgb}{0.56,0.27,0.68}
\definecolor{red}{rgb}{0.95,0.4,0.4}
\definecolor{purered}{rgb}{1,0,0}
\definecolor{blue}{rgb}{0.4,0.4,0.95}
\definecolor{darkblue}{rgb}{0,0,0.8}
\definecolor{grey}{rgb}{0.6,0.6,0.6}
\definecolor{col1}{RGB}{232, 161, 148}
\definecolor{col2}{RGB}{148, 187, 232}
\definecolor{col3}{RGB}{206, 239, 255}
\definecolor{lightgrey}{rgb}{0.85,0.85,0.85}
\definecolor{lightlightgrey}{rgb}{0.9,0.9,0.9}
\definecolor{verylightBG}{rgb}{0.9,0.99,0.99}
\definecolor{darkgreen}{rgb}{0.3, 0.75, 0.3}

\definecolor{cvprblue}{rgb}{0.21,0.49,0.74}
\usepackage[pagebackref,breaklinks,colorlinks,citecolor=cvprblue]{hyperref}

\definecolor{purple}{rgb}{0.56,0.27,0.68}
\definecolor{red}{rgb}{0.95,0.4,0.4}
\definecolor{purered}{rgb}{1,0,0}
\definecolor{blue}{rgb}{0.4,0.4,0.95}
\definecolor{darkblue}{rgb}{0,0,0.8}
\definecolor{lightblue}{RGB}{170,180,255}
\definecolor{grey}{rgb}{0.6,0.6,0.6}
\definecolor{col1}{RGB}{232, 161, 148}
\definecolor{col2}{RGB}{148, 187, 232}
\definecolor{col3}{RGB}{206, 239, 255}
\definecolor{lightgrey}{rgb}{0.85,0.85,0.85}
\definecolor{lightlightgrey}{rgb}{0.9,0.9,0.9}
\definecolor{verylightBG}{rgb}{0.9,0.99,0.99}
\definecolor{darkgreen}{rgb}{0.3, 0.75, 0.3}
\definecolor{darkgreen}{RGB}{54, 185, 61}

\usepackage[utf8]{inputenc} 
\usepackage[T1]{fontenc}    
\usepackage{hyperref}       
\usepackage{url}            
\usepackage{booktabs}       
\usepackage{amsfonts}       
\usepackage{nicefrac}       
\usepackage{microtype}      
\usepackage{colortbl}
\usepackage[dvipsnames]{xcolor}
\usepackage{graphicx}
\usepackage{multirow}
\usepackage{caption}
\usepackage{makecell}
\usepackage{arydshln}
\usepackage{array, comment}
\usepackage{nicematrix}
\usepackage{sidecap}

\def\paperID{1761} 
\def\confName{CVPR}
\def\confYear{2025}

\title{Solving Instance Detection from an Open-World Perspective}

\author{
  Qianqian Shen$^{1,}$\thanks{Equal contributions.}\quad Yunhan Zhao$^{2,*}$\quad Nahyun Kwon$^{3}$\quad Jeeeun Kim$^{3}$\quad Yanan Li$^{4}$\quad Shu Kong$^{5}$ \\
  {\small $^1$Zhejiang University, $^2$UC Irvine, $^3$Texas A\&M University, $^4$Zhejiang Lab} \\
  {\small $^5$University of Macau,  $^6$Institute of Collaborative Innovation} \\ 
  {\small \em website and code: \url{https://shenqq377.github.io/IDOW}}
}

\begin{document}
\maketitle

\begin{abstract}

Instance detection (InsDet) aims to localize specific object instances within a novel scene imagery based on given visual references. Technically, it requires proposal detection to identify all possible object instances, followed by instance-level matching to pinpoint the ones of interest. Its open-world nature supports its broad applications from robotics to AR/VR but also presents significant challenges: methods must generalize to unknown testing data distributions because (1) the testing scene imagery is unseen during training, and (2) there are domain gaps between visual references and detected proposals. Existing methods tackle these challenges by synthesizing diverse training examples or utilizing off-the-shelf foundation models (FMs). However, they only partially capitalize the available open-world information. In contrast, we approach InsDet from an Open-World perspective, introducing our method IDOW. We find that, while pretrained FMs yield high recall in instance detection, they are not specifically optimized for instance-level feature matching. Therefore, we adapt pretrained FMs for improved instance-level matching using open-world data. Our approach incorporates metric learning along with novel data augmentations, which sample distractors as negative examples and synthesize novel-view instances to enrich the visual references. Extensive experiments demonstrate that our method significantly outperforms prior works, achieving $>$10 AP over previous results on two recently released challenging benchmark datasets in both conventional and novel instance detection settings.

\end{abstract}

\section{Introduction}
\label{sec:intro}

\begin{figure}[t]
\centering
\includegraphics[width=0.99\linewidth]{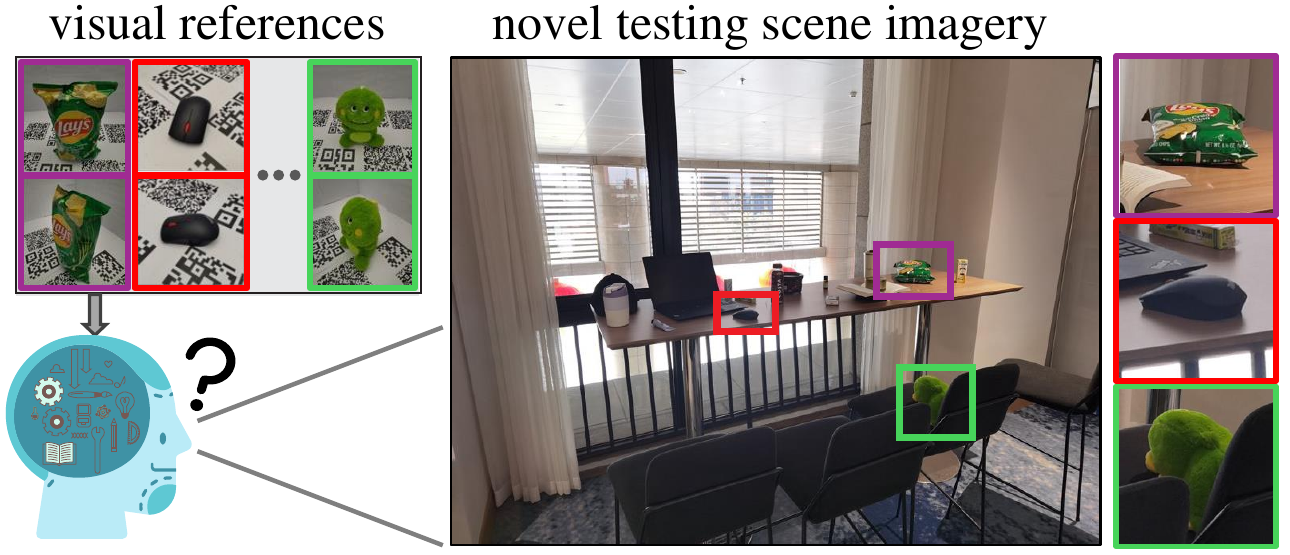}
\vspace{-4mm}
\caption{\small 
Instance Detection (InsDet) aims to localize specific object instances in novel scene imagery based on given visual references (\emph{aka} support templates).
It is a challenging problem due to its open-world nature:
the testing scene images are unseen during training and thus are unknown to InsDet models,
and visual references and detected proposals have domain gaps (e.g., due to occlusions and arbitrary lighting conditions in the latter).
}
\vspace{-5mm}
\label{fig:InsDet_illustration}
\end{figure}

Instance detection (InsDet) aims to localize object instances of interest in novel scene imagery  (Fig.~\ref{fig:InsDet_illustration}), where these objects are specified by some visual references (as known as support templates)~\cite{tang2017multiple,mercier2021deep,bormann2021real, shen2023high, li2024voxdet}.
Its open-world nature supports its wide-ranging applications in robotics and AR/VR.
For example, a robot can be commanded to search for a customer's suitcase at an airport~\cite{li2024voxdet}, or \emph{my-keys} in my cluttered bedroom~\cite{zhao2024instance}.
Its open-world nature also presents significant challenges: 
during training, one has no knowledge of testing data distribution, which can be instantiated by the unknown scene imagery~\cite{dwibedi2017cut}, novel object instances encountered only in testing~\cite{li2024voxdet}, and domain gaps between visual references and detected proposals~\cite{shen2023high}.

\begin{figure*}[t]
\centering
\small
{\scriptsize \ 
{\bf (a)} background imagery~\cite{Georgakis2017SynthesizingTD, dwibedi2017cut}
\hspace{3mm}
{\bf (b)} object images~\cite{li2024voxdet} 
\hspace{2mm} 
{\bf (c)} Foundation Models~\cite{shen2023high} 
\hspace{12mm}
{\bf (d)} Solving Instance Detection in the Open World ({\bf ours})}
\hspace{20mm} \
\\
\includegraphics[width=0.99\linewidth, trim={0 0 0 7mm},clip]{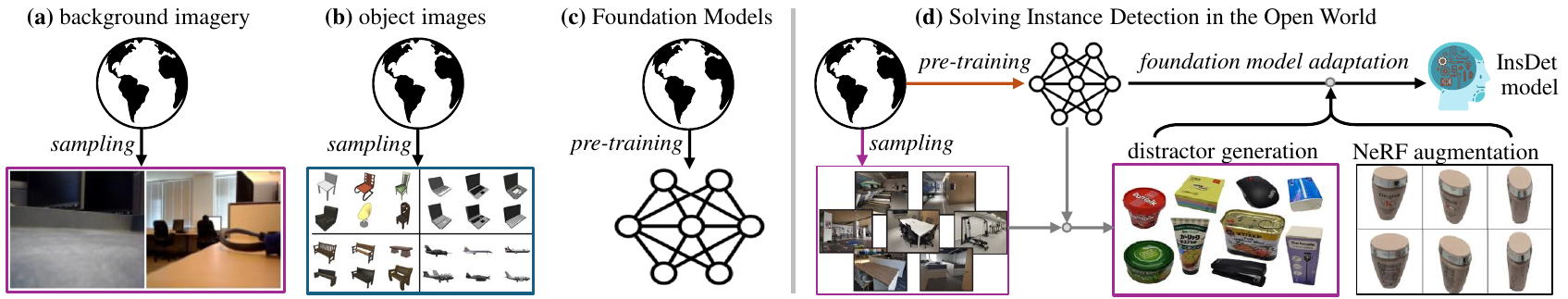}
\vspace{-2mm}
\caption{\small 
Existing InsDet methods leverage the open-world information in different aspects, such as {\bf (a)} background image sampling (from the open world) to synthesize training data~\cite{Georgakis2017SynthesizingTD, dwibedi2017cut},
{\bf (b)} object image sampling (from the open world) to learn feature representations~\cite{li2024voxdet},
and {\bf (c)} foundation model utilization (pretrained in the open world) for proposal detection and instance-level feature matching~\cite{shen2023high}.
{\bf (d)} As FMs are not specifically designed for instance-level feature matching required by InsDet, we propose to adapt them by leveraging rich data sampled from the open world.
We gather data from multiple sources:
(1) any available visual references of instances in the CID setting,
(2) abundant multi-view object images sampled in the open world similar to  {\bf (b)},
(3) synthetic data by training  NeRF~\cite{mildenhall2021nerf, barron2023zip} to generate novel-view images based on the given instances,
(4) \emph{distractors} by running FMs (esp. SAM~\cite{kirillov2023segment}) on random open-world imagery to generate random object-like proposals (Fig.~\ref{fig:NeRF-vis}).
We use the data above to adapt FM through metric learning. The technical novelty of our work lies in (3) and (4), as well as  the design choice of metric learning to adapt FMs for InsDet.
}
\vspace{-3mm}
\label{fig:open-world-practice}
\end{figure*}

{\bf Status quo.}
The literature of InsDet has two settings and existing InsDet  methods attempt to address the open world in different aspects (Fig.~\ref{fig:open-world-practice}).
The \emph{conventional instance detection (CID)} setting provides visual references of object instances for training.
Yet, the testing scene is still unknown in training time.
Therefore, early method Cut Paste Learn (CPL) collects random background images to model diverse testing scenes (Fig.~\ref{fig:open-world-practice}a), cuts instances from the visual references, and pastes them on these background images to construct a synthetic training set~\cite{Georgakis2017SynthesizingTD, dwibedi2017cut}.
It is worth noting that background image sampling is utilizing open-world information in a way of data collection in the open world.
On the other hand, the \emph{novel instance detection (NID)} setting, introduced recently~\cite{li2024voxdet}, studies detecting novel object instances that are specified only in testing time where a trained model is not allowed to be tuned further.
To approach NID, VoxDet~\cite{li2024voxdet} uses large-scale synthetic data (Fig.~\ref{fig:open-world-practice}b) to learn 3D voxel representations from visual references and a matching function between 2D proposals and 3D voxels.
Recent work OTS-FM \cite{shen2023high} uses off-the-shelf pretrained foundation models (Fig.~\ref{fig:open-world-practice}c) for InsDet.
Specifically, it uses SAM~\cite{kirillov2023segment} to detect object proposals in scene imagery, and DINOv2~\cite{oquab2023dinov2} to represent visual references and proposals for instance-level feature matching.
OTS-FM is a non-learned method and hence is applicable in both CID and NID settings. Importantly, it achieves the state-of-the-art performance on the recently released benchmark  HR-InsDet~\cite{shen2023high}.

{\bf The open-world perspective.}
As illustrated in Fig.~\ref{fig:InsDet_illustration}, 
the challenges of InsDet lie in its open-world nature: the unknown testing data distribution, and domain gaps between visual references and detected proposals.
To address the open-world challenges, existing methods take different strategies to exploit the open-world information (Fig.~\ref{fig:open-world-practice}a-c). 
For example,
to address the unknown testing scene imagery, some methods sample diverse background images~\cite{Georgakis2017SynthesizingTD, dwibedi2017cut}, demonstrating a practice of sampling data in the open world.
To mitigate the domain gap between detected proposals and visual references,
\cite{li2024voxdet} attempts to learn features for instance-level matching by utilizing external datasets (ShapeNet~\cite{chang2015shapenet} and ABO~\cite{collins2022abo}). This data utilization can be considered as data collection in the open world.
For better proposal detection and more discriminative feature representations, \cite{shen2023high} adopts off-the-shelf pretrained foundation models (FMs) SAM~\cite{kirillov2023segment} 
and DINOv2~\cite{oquab2023dinov2}, respectively.
One can think of FMs as being pretrained in the open world.
Different from existing methods which have not fully capitalized the open-world information,
we take the open-world perspective as the first principle and solve \emph{InsDet} by exploiting the \emph{Open World} (Fig.~\ref{fig:open-world-practice}d).  We call our approach \emph{IDOW}.

{\bf Insights and novelties.}
Our insights are rooted in the understanding of InsDet's open-world nature and the practice of existing methods.
Existing methods address open-world issues to some extent (Fig.~\ref{fig:open-world-practice}), particularly by sampling data in the open world~\cite{dwibedi2017cut, kong2021opengan, li2024voxdet} and adopting foundation models (FMs) pretrained in the open world~\cite{shen2023high}.
These motivate us to develop InsDet methods in the open world (\emph{IDOW}). 
We find that open-world detectors yield high recall on object instances (see precision-recall curves in Fig.~\ref{fig:pr_curve} of the Appendix) but FMs are not tailored to instance-level matching between proposals and visual references -- directly using an FM for InsDet is sub-optimal. 
Therefore, we adapt FMs to instance-level feature representation by metric learning.
We propose novel and simple techniques for data augmentation to enhance FM adaptation and feature learning: distractor sampling, and multi-view synthesis.
The former helps learn more discriminative features for instance comparison, and the latter enhances visual references particularly when they are few.
With these techniques, our IDOW achieves significant boosts over previously reported results on recently released challenging benchmarks (Table \ref{tab:insdet-CID} and \ref{tab:robotools-NID}).

{\bf Contributions.}
We make three major contributions:
\begin{enumerate}
    \item 
    We elaborate on the open-world challenge of InsDet and motivate the necessity of developing  methods in the open world. We solve InsDet by exploiting diverse \emph{open data} available in the open world and FMs pretrained therein.
    
    \item
    We introduce simple and effective techniques to adapt FMs to InsDet: a metric learning loss, and data augmentation with distractor sampling and novel-view synthesis.

    \item
    We extensively compare our approach with existing ones on two recently released datasets, demonstrating that our approach outperforms them by $>$10 AP in both conventional and novel instance detection settings.
\end{enumerate}

\section{Related Work}
\label{sec:related-work}
{\bf Instance detection.}
Early methods use local features such as SIFT~\cite{lowe2004distinctive} and SURF~\cite{bay2006surf} to match visual references~\cite{quadros2012occlusion} and image regions to localize instances~\cite{hinterstoisser2011gradient}. 
Recent methods train deep neural networks for instance-agnostic proposal detection and proposal-instance matching, achieving significant improvements~\cite{mercier2021deep, li2024voxdet, shen2023high}.
In the \emph{conventional instance detection} (CID) setting, 
visual references of object instances are given during training.
To approach CID, prevalent method synthesize training data in a cut-paste-learn (CPL) strategy~\cite{Georgakis2017SynthesizingTD, dwibedi2017cut}: cutting instances from visual references and pasting them on random background images, then learning a detector on such synthetic data.
Other works strive to obtain more training samples by rendering realistic instance examples~\cite{kehl2017ssd, hodavn2019photorealistic}, using data augmentation~\cite{dwibedi2017cut} and synthesizing better training images~\cite{Lai2014UnsupervisedFL, dwibedi2017cut, georgakis2017synthesizing}.
Moreover, recent literature introduces the \emph{novel instance detection} (NID) setting~\cite{li2024voxdet}, which requires detecting novel instances specified \emph{only} in testing and does not allow for further model finetuning during testing.
VoxDet \cite{li2024voxdet} utilizes a pretrained open-world detector~\cite{kim2022learning} for proposal detection.
Nowadays, the open-world perspective advances proposal detection in a way of learning more generalizable detectors and carrying out open-vocabulary recognition~\cite{liu2023grounding}.
While a foundational open-vocabulary detector can attach a textual description (e.g., a short phrase) to each proposal, simply using it is insufficient for instance-level matching.
\cite{shen2023high} realizes this issue although foundational detectors can yield nearly perfect recall.
As a result, the recent method OTS-FM uses another FM DINOv2~\cite{oquab2023dinov2} for proposal-instance matching~\cite{shen2023high}.
Differently, we solve InsDet from the open-world perspective, exploiting open FMs and open data to learn more discriminative instance-level feature representations.

{\bf Foundation models} (FMs) are pre-trained on web-scale data, which can be thought of as being sampled in the \emph{open world}.
Different FMs are trained on different types of data and can shine in different downstream tasks related to visual perception, natural language processing, or both.
As this work concerns about visual perception, we briefly review Vision-Language Models (VLMs) and Visual Foundation Models (VFMs).
VLMs are pretrained on web-scale image-text paired data~\cite{radford2021learning, jia2021scaling, liu2023llava, liu2023grounding}, demonstrating impressive results in high-level tasks such as visual grounding and image captioning.
VFMs are pretrained primarily on visual data~\cite{chen2020simple, caron2021emerging, zhou2021ibot, he2022vlmae, oquab2023dinov2,  touvron2022deit, kirillov2023segment, wang2024segment} and can yield impressive visual perception results such as proposal or open-set object detection~\cite{zhang2022dino, liu2023grounding}.
In InsDet, the mainstream framework requires proposal detection and visual feature matching.
In this framework, open-world detectors appear to achieve high recall on the instances of interest, owing to that the open-world pretrained detectors generalize quite well in the open world~\cite{zhang2022dino, liu2023grounding, ren2024grounded}.
In contrast, feature matching still remains a challenge in improving InsDet.
Therefore, we focus on improving features for InsDet by adapting VFMs using data sampled in the open world.
\cite{shen2023high} shows that using an off-the-shelf FM DINOv2~\cite{oquab2023dinov2} for feature matching greatly enhances InsDet detection; yet, FM adaptation has been still under-explored in InsDet, although it  is a well studied in other areas such as few-shot perception~\cite{clap24, madan2024revisiting, liu2025few} and zero-shot perception~\cite{goyal2023finetune, Parashar2023prompting,parashar2024neglected}. 
Our work introduces simple and novel techniques to adapt FMs to improve InsDet.

\begin{figure*}[t]
\centering
\small
\hspace{-12mm} Support \hspace{11mm} Ground Truth \hspace{18mm} Cut-Paste-Learn~\cite{dwibedi2017cut} \hspace{18mm} OTS-FM~\cite{shen2023high} \hspace{22mm} {\bf IDOW (ours)} \hspace{12mm} 
\\
\includegraphics[width=0.99\textwidth, trim={0 0 0 0},clip]{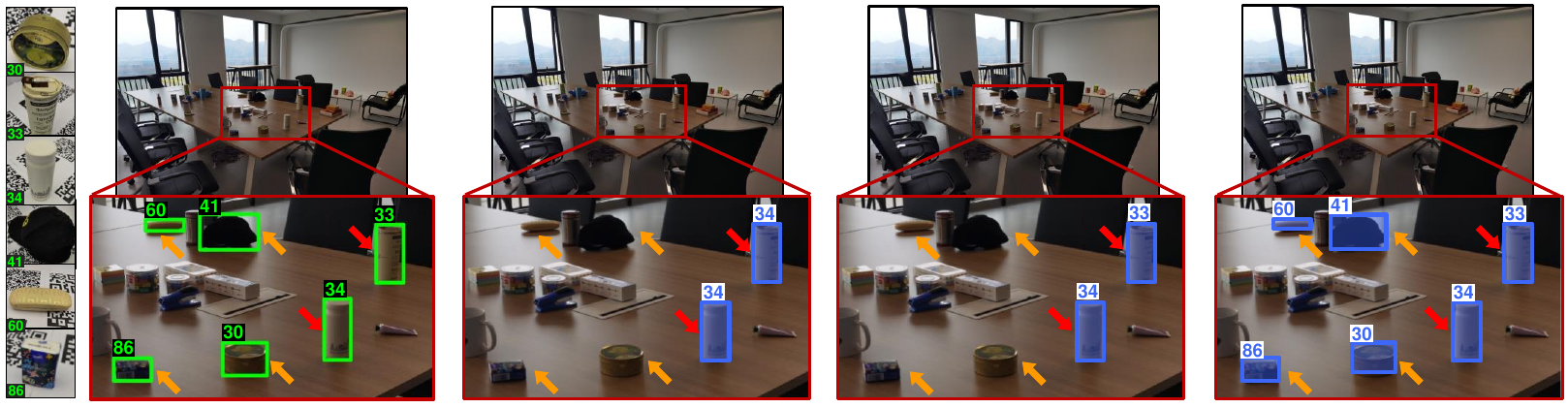}
\vspace{-3mm}
\caption{\small 
{\bf Visual comparison of InsDet results by different methods in the CID setting} on HR-InsDet~\cite{shen2023high}.
The testing scene image contains sparse placement of small instances with challenging illumination.
We mark the ground-truth and predictions using \textcolor{darkgreen}{green} and 
{\setlength{\fboxsep}{1pt}\colorbox{lightblue}{blue}} boxes, respectively.
We attach instance IDs to them to highlight whether the instance recognition is correct compared to the visual references (i.e., the leftmost references).
Compared with Cut-Paste-Learn 
and OTS-FM, our IDOW detects more instances (see \textcolor{orange}{orange arrows}) with better accuracy (see \textcolor{purered}{red arrows}).
Comparison between OTS-FM and IDOW suggests suggests that adapting the FM DINOv2 yields better features for IDOW to perform much better in front of domain shifts, i.e., challenging illumination conditions in this testing scene image.
}
\label{fig:closed-world-vis}
\vspace{-3mm}
\end{figure*}

{\bf The open-world data} is publicly available for pretraining FMs.
Coping with such data is also often challenging.
For example, data between training and testing has distribution shifts or domain gaps~\cite{shi2024lca, liu2025few},
as clearly instantiated in InsDet by the unknown testing scene imagery, in addition to the domain gaps between (clean) visual references and (occluded) proposals (Fig.~\ref{fig:InsDet_illustration}).
Speaking of distribution shifts, some existing methods sample open-world data, demonstrating a need to develop robust models in the open world~\cite{hendrycks2018deep, kong2021opengan}.
Existing InsDet methods also sample open-world data such as diverse background images and instance examples (Fig.~\ref{fig:open-world-practice})~\cite{Georgakis2017SynthesizingTD, dwibedi2017cut}.
Even so, while open-world data is abundant, most of it is not directly useful for training models for specific downstream tasks.
In contrast,
data augmentation aims to synthesize more tailored data to enhance models' robustness and generalization, including geometry-aware augmentation (e.g., random cropping and flipping)~\cite{krizhevsky2012imagenet, chen2020simple, zhao2021camera}, photometric-aware augmentation (e.g., color jittering)~\cite{volpi2018adversarial, kim2020data, volpi2018generalizing}, 
and robustness-aware augmentation (e.g., adversarial perturbation)~\cite{zhang2019adversarial}. 
It is worth noting that 
Neural Radiance Field (NeRF)~\cite{NeRF}, a recent rendering technique, can be used to synthesize novel-view images.
In this work, we particularly adopt NeRF to synthesize novel-view visual references of object instances and use them to adapt FMs towards more tailored features to InsDet.
To the best of our knowledge, using NeRF is under-explored in the literature of InsDet, but we find that doing so remarkably improves InsDet performance.

\section{Instance Detection: Protocols and Methods}
\label{sec:method}

We first introduce the protocols of InsDet, including the problem definition, evaluation, and two settings.
We then present our techniques to enhance foundation model adaptation to InsDet.

\subsection{Protocols}

{\bf Problem definition.}
Instance Detection (InsDet) aims to detect specific object instances from a 2D scene image (Fig.~\ref{fig:InsDet_illustration}), where the objects of interest are  defined by a ``support set'' of some visual references~\cite{dwibedi2017cut, shen2023high, li2024voxdet}.
Previous works capture the visual references from multiple camera views, with  QR code being pasted to help estimate camera poses (see visual references in Fig.~\ref{fig:InsDet_illustration}).

{\bf Challenges.}
InsDet is a challenging problem due to its open-world nature:
one has zero knowledge about the testing data distribution which is instantiated by unknown testing scenes, never-before-seen matters, arbitrary occlusions, background clutters, etc.
In particular, object instances of interest can be novel and are only defined during testing, as emphasized by the novel instance detection setting below. Nevertheless, InsDet models must be capable of detecting \emph{any} object instances of interest and robustly comparing them with visual references.

{\bf Two settings.}
There are two settings of InsDet, which differ in whether object instances are pre-defined during training or defined on the fly in testing time. 
\begin{itemize} 
\item {\em Conventional Instance Detection (CID).} 
    In this setting,
    all object instances of interest are pre-defined during training~\cite{dwibedi2017cut}.
    That said, one can use their visual references as training data to train detectors. Yet, only visual references are available and testing scene data is still unknown. 
    Hence, InsDet methods in CID often choose to sample diverse background images in the open world~\cite{dwibedi2017cut, georgakis2017synthesizing}.
    This setting simulates application scenarios in AR/VR.
    For instance, robots should memorize the items concerned by the customer so as to provide better customized services.
\item {\em Novel Instance Detection (NID).}
    This setting requires InsDet models to detect \emph{novel} object instances defined only in testing time and does not allow the models to be finetuned further during testing~\cite{li2024voxdet}.
    In this setting, the \emph{de facto} practice to train InsDet models turns to external data, which can be either synthetic data or those sampled in the open world.
    This setting simulates some real-world scenarios, for instance, robots must search for a never-before-seen luggage of a customer at an airport.
\end{itemize}

\subsection{Learning Instance Detector in Open World}
We present our solution of instance detection from an open-world perspective (IDOW), focusing on foundation model adaptation and data augmentation.

\subsubsection{Foundation Model Adaptation}

As an open-world detector yields quite high recall~\cite{shen2023high, li2024voxdet},
we focus on adapting a foundation model to enhance feature representations for InsDet.
Although a foundational feature model (e.g., DINOv2~\cite{oquab2023dinov2}) offers feature representations applicable to various downstream tasks, it is not tailored to InsDet.
To improve feature representation, one can finetune this FM over relevant data, such as those sampled in the open world \cite{li2024voxdet}.
Intuitively, the finetuned feature is expected to match proposals and visual references better if they are from the same object instances; otherwise, distinguish them.
Therefore, we propose to use a metric learning loss for FM adaptation.
Moreover, we sample \emph{distractors} from imagery in the open world. The distractors serve as negatives in metric learning.

\begin{figure*}[t]
\centering
\small
\hspace{-5mm} Support \hspace{15mm} Ground Truth \hspace{19mm} VoxDet~\cite{li2024voxdet} \hspace{19mm} OTS-FM~\cite{shen2023high} \hspace{18mm} {\bf IDOW (ours)} \hspace{20mm} 
\\
\includegraphics[width=0.99\textwidth, trim={0 0 0 0},clip]{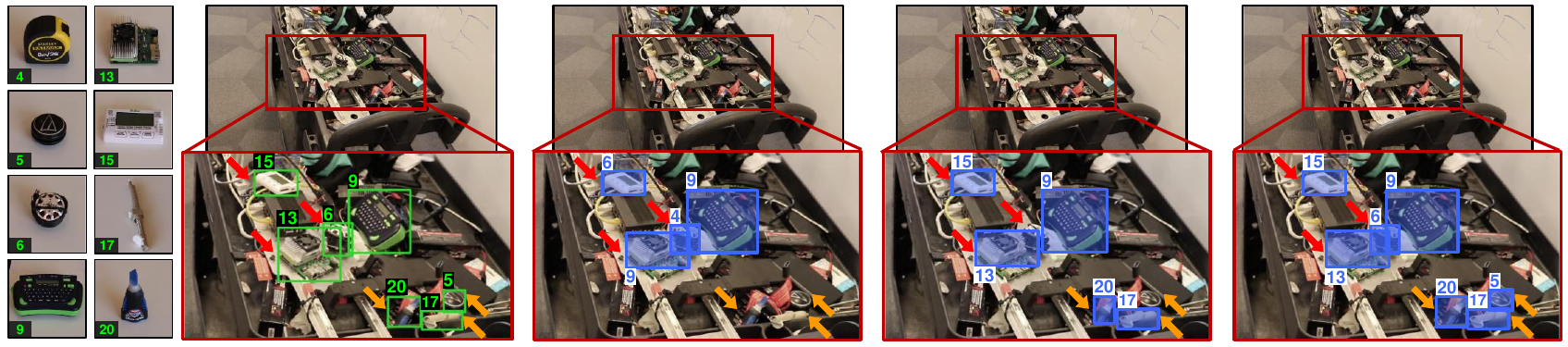}
\vspace{-3mm}
\caption{\small 
{\bf Visual comparison of InsDet results by different methods in the NID setting} on RoboTools~\cite{li2024voxdet}.
On the cluttered testing scene image, we mark the ground-truth and predictions using \textcolor{darkgreen}{green} and 
{\setlength{\fboxsep}{1pt}\colorbox{lightblue}{blue}}
boxes, respectively. 
We attach instance IDs to them to highlight whether the instance recognition is correct compared to the visual references
(i.e., the leftmost references). 
Compared with VoxDet and OTS-FM, our IDOW detects more instances (see \textcolor{orange}{orange arrows}) with better accuracy (see \textcolor{purered}{red arrows}).
Comparing OTS-FM and our IDOW suggests that, although having not seen the visual references during adaption in this NID setting, features by our IDOW are a better representation for InsDet than the off-the-shelf foundation model (i.e., DINOv2) used in OTS-FM.
}
\vspace{-3mm}
\label{fig:open-world_RoboTools}
\end{figure*}

We denote an FM as $f_{\theta}$, parameterized by $\theta$. This FM serves as the feature extractor $f_{\theta}(\I): \I \rightarrow \mathbb{R}^q$ 
to produce $q$-dimensional features for any given visual reference image or proposal $\I$.
We finetune $f_{\theta}(\cdot)$ to better serve InsDet such that features of reference images from the same instance should present high similarity, otherwise low similarity.
We achieve this by finetuning $f_{\theta}$ using a simple metric learning loss $\ell$.
Specifically,
we construct triplets of training data $(\I_a, \I_p, \I_n)$, where $\I_a$, $\I_p$ and $\I_n$ represent an anchor visual reference, positive and negative samples, respectively.
The training data contains examples sampled or synthesized in the open world~\cite{li2024voxdet}, as well as visual references available in the CID setting.
Negative examples are cross-instance visual examples and distractors (detailed later).
Below is the loss:
\vspace{-1mm}
\begin{align}\small
    \ell = \Big[
    d\big(f_{\theta}(\I_a), f_{\theta}(\I_p) \big) - 
    d\big(f_{\theta}(\I_a), f_{\theta}(\I_n)\big) +
    \alpha \Big]_{+},
\end{align}
where $\alpha$ is a hyper-parameter determining the margin between an instance and other negative data.
$d(\cdot)$ measures the distance between two examples, e.g., inverse cosine similarity used in our work.
After training, we use the learned features to represent both proposals and visual references.
We follow the instance-proposal matching pipeline~\cite{shen2023high} to produce the final InsDet results: computing pairwise similarities between each proposal and each visual reference,
running stable matching algorithm~\cite{gale1962college, mcvitie1971stable},
and returning the matched pairs if they have similarity scores greater than a predefined threshold (0.4 as used in \cite{shen2023high}).

{\bf Hard example sampling.}
To construct training batches of triplets, it is important to effectively sample negative examples, easy negative examples do not provide sufficient gradients during training \cite{shrivastava2016training}.
In this work, we adopt a batch-level hard negative sampling strategy. 
In each training batch, we first sample a reference image as an anchor $\I_a$. 
We then sample a random example from the same instance as the positive one $\I_p$.
We finally sample a hard negative example from the union of the reference images of all the other instances and the distractors  $S_{dt}$ (explained in the next subsection).
Mathematically, we find hard negative patch $\I_{hn}$ with the objective: $\arg\min_{\I_n} d(f_{\theta}(\I_a), f_{\theta}(\I_n)),\ \I_n \in \{C(\I_i) \neq C(\I_a), \text{for} \ i=1 \dots N \} \cup S_{dt}$, where $C(\I_i)$ represents the instance ID of the data $\I_i$, and $N$ indicates the total number of visual reference images.

\subsubsection{The Proposed Data Augmentation}
\label{ssec:augmentation}

To better adapt FM for instance-level feature matching,
we introduce two data augmentation techniques below.

{\bf Distractor sampling} aims to sample patches of random background images as universal negative data to all object instances. 
It is a data augmentation technique, expected to define the open space and help features better characterize meaningful object instances.
Distractor sampling has been adopted in the literature. 
For example, open-set recognition samples distractors as a separate class, known as ``the-other-class'' or ``background-class'', to define the open space.
In InsDet, the cut-paste-learn strategy~\cite{dwibedi2017cut} samples random background photos to define the background content.
In this work, we run SAM~\cite{kirillov2023segment} on random background photos and use the segments as distractors.
One might worry about the possibility of sampled distractors being exactly some instances of interest and including them as negatives may destabilize model adaptation. 
Yet, given a large number of distractors, this rarely becomes an issue as demonstrated by the success of various self-supervised learning methods~\cite{oquab2023dinov2} which may sample examples serving as both positive and negative data at the same time.

\begin{figure}[t]
\centering
\small
\includegraphics[width=0.99\linewidth, trim={0 0 0 0},clip]{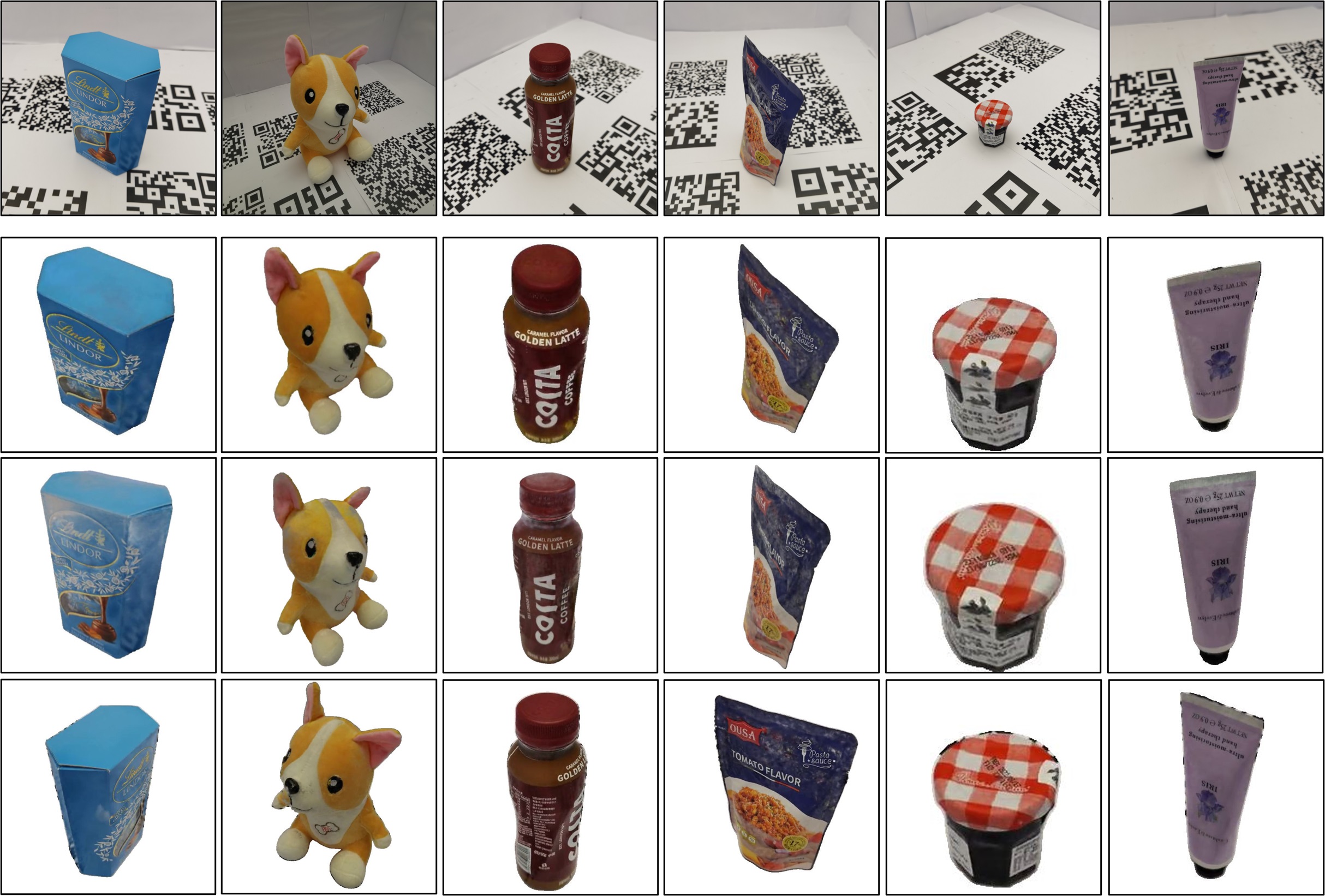}
\vspace{-3mm}
\caption{\small 
We use NeRF to synthesize novel-view object images to augment the limited given visual references.
In this work, we train per-instance Zip-NeRF~\cite{barron2023zip}.
We visualize synthesized images at different angles (in row 2-4) together with the raw visual references (in the first row).
The visual references are from HR-InsDet dataset~\cite{shen2023high}, where the QR code is used for camera pose estimation.
Overall, we find synthesized images show high visual quality.
}
\label{fig:NeRF-vis}
\vspace{-5mm}
\end{figure}

{
\setlength{\tabcolsep}{0.1em}  
\begin{table*}[t]
\centering
\small
\caption{\small
{\bf Benchmarking results in the CID setting} on the HR-InsDet dataset. We compare our IDOW with state-of-the-arts and make three salient conclusions. First, IDOW significantly outperforms previous methods, e.g., IDOW$_{\text{GroundingDINO}}$ (57.01 AP) $>$ OTS-FM$_{\text{GroundingDINO}}$ (51.68 AP) $>$ CPL$_{\text{DINO}}$ (27.99 AP). 
This confirms the importance of addressing InsDet from the open-world perspective.
Second, adapting FMs by our IDOW further boosts the performance by 5-7 AP, e.g., IDOW$_{\text{SAM}}$ (48.75 AP) $>$ OTS-FM$_{\text{SAM}}$ (41.61 AP). 
Third, IDOW and OTS-FM are applicable to different pretrained FMs and adopting stronger FMs achieves better performance, e.g., using GroundingDINO yields $>$8 AP than SAM in IDOW. 
}
\vspace{-3mm}
\begin{tabular*}{\hsize}{@{}@{\extracolsep{\fill}}llcccccccc@{ }}
\toprule
{\bf Method} & \textcolor{grey}{\bf Venue \& Year} & \multicolumn{6}{c}{\bf AP} & {\bf AP$_{50}$}  & {\bf AP$_{75}$} \\ 
\cmidrule(l){3-8} 
 & & {\tt avg} & {\tt hard} & {\tt easy} & {\tt small} & {\tt medium} & {\tt large} & &\\
\midrule
CPL$_{\text{FasterRCNN}}$~\cite{NIPS2015_fasterRCNN, dwibedi2017cut} & \textcolor{grey}{NeurIPS 2015}
& \cellcolor{col3}19.54 & 10.26 & 23.75
& 5.03 & 22.20 & 37.97 
& 29.21 & 23.26 
\\
CPL$_{\text{RetinaNet}}$~\cite{lin2017focal, dwibedi2017cut} & \textcolor{grey}{ICCV 2017}
& \cellcolor{col3}22.22 & 14.92 & 26.49 
& 5.48 & 25.80 & 42.71 
& 31.19 & 24.98 
\\
CPL$_{\text{CenterNet}}$~\cite{dwibedi2017cut, zhou2019objects} & \textcolor{grey}{CVPR 2019}
& \cellcolor{col3}21.12 & 11.85 & 25.70 
& 5.90 & 24.15 & 40.38
& 32.72 & 23.60 
\\
CPL$_{\text{FCOS}}$~\cite{dwibedi2017cut, tian2019fcos} & \textcolor{grey}{ICCV 2019}
& \cellcolor{col3}22.40 & 13.22 & 28.68
& 6.17 & 26.46 & 38.13 
& 32.80 & 25.47 
\\
CPL$_{\text{DINO}}$~\cite{dwibedi2017cut, zhang2022dino} & \textcolor{grey}{ICLR 2023}
& \cellcolor{col3}27.99 & 17.89 & 32.65 
& 11.51 & 31.60 & 48.35 
& 39.62 & 32.19  
\\
OTS-FM$_{\text{SAM}}$~\cite{shen2023high, kirillov2023segment} & \textcolor{grey}{NeurIPS 2023}
& \cellcolor{col3}41.61 & 28.03 & 47.57 
& 14.58 & 45.83 & 69.14 
& 49.10 & 45.95 \\
OTS-FM$_{\text{GroundingDINO}}$~\cite{shen2023high, liu2023grounding}  & \textcolor{grey}{NeurIPS 2023} & \cellcolor{col3}51.68 & 37.23 & 58.72
& 28.79 & 58.55 & 69.22
& 62.50 & 56.78
\\ 
\midrule
IDOW$_{\text{SAM}}$ & 
& \cellcolor{col3}48.75 & 32.09 & 56.50
& 20.75 & 55.26 & 73.43
& 57.59 & 54.06 
\\
IDOW$_{\text{GroundingDINO}}$   & 
& \cellcolor{col3}\bf{57.01} & \bf{40.74} & \bf{64.36}
& \bf{35.25} & \bf{62.98} & \bf{73.64} 
& \bf{69.33} & \bf{62.84} 
\\
\bottomrule
\end{tabular*}
\vspace{-4mm}
\label{tab:insdet-CID}
\end{table*}
}

{\bf Novel-view synthesis.}
Each object instance might be represented by only a few visual examples, especially in the CID setting which provides limited real visual references for instances of interest.
To augment available visual references for better adapting FM,
we learn a NeRF~\cite{NeRF} for each instance and use NeRF to synthesize novel-view images.
In this process, we estimate relative camera poses (required by NeRF) using COLMAP~\cite{schoenberger2016sfm}, as commonly done in NeRF methods~\cite{NeRF,barron2023zip} and InsDet approaches~\cite{li2024voxdet,shen2023high}.
Fig.~\ref{fig:NeRF-vis}  displays some random NeRF-generated novel-view images with good quality in visuals.
It is worth noting that a NeRF needs to be trained only once on the visual references of an object instance. Once it is trained, we use it to generate more visual references not only for training but for testing.
In particular, such NeRF-synthesized novel-view images can be stored together with the original visual references in testing for proposal-instance matching.
Doing so significantly improves InsDet performance (Table~\ref{tab:ablation} and Fig.~\ref{fig:ablation-num_views}).
To the best of our knowledge, our work is the first that exploits NeRF to synthesize data for InsDet.

\section{Experiments}
\label{sec:experiments}
In this section, we first introduce the experimental setup, then evaluate our IDOW by comparing existing methods in both CID and NID settings, and finally conduct ablation studies to analyze our approach.

\subsection{Experimental Setup}

{\bf Datasets.}
We use two recently published datasets in our experiments.
HR-InsDet~\cite{shen2023high} is developed for the CID setting, containing 100 daily object instances, 160 high-resolution testing images from 14 indoor scenarios (see Fig.~\ref{fig:closed-world-vis} for an example).
RoboTools~\cite{li2024voxdet} is developed for the NID setting, containing 20 robotic tool instances and 1,581 testing images from 24 indoor scenarios (see Fig.~\ref{fig:open-world_RoboTools} for an example). 
For a fair comparison, we use the OWID dataset released by \cite{li2024voxdet} to adapt FM in the NID setting.

{\bf Metrics.}
We use metrics adopted in both datasets mentioned above,
including average precision (AP) at IoU thresholds from 0.5 to 0.95 with the step size 0.05 as the primary metric and also report AP$_{50}$ and AP$_{75}$ averaged over all instances with IoU threshold as 0.5 and 0.75, respectively. 
The HR-InsDet dataset contains tags for hard/easy testing scenes and small/medium/large object instances. We follow~\cite{shen2023high} to break down results on these tags.

{\bf Compared methods.}
We compare previous methods including CPL, OTS-FM, and VoxDet.
These methods have options to use different components (so do ours).
For example, CPL can train a detector using one of the popular architectures Faster RCNN~\cite{NIPS2015_fasterRCNN},
RetinaNet~\cite{lin2017focal}, CenterNet~\cite{zhou2019objects}, FCOS~\cite{tian2019fcos}, and DINO~\cite{zhang2022dino}.
OTS-FM can use SAM or GroundingDINO for proposal detection.
We implement these methods by using different backbones to improve their performance (Table~\ref{tab:insdet-CID}).
We use subscripts to mark the backbones used in each method.
Moreover, we compare more methods in the NID setting, including OS2D~\cite{osokin2020os2d}, DTOID~\cite{mercier2021deep}, and OLN$_\text{Corr.}$~\cite{kim2022learning}.

\begin{figure*}[t]
\centering
\includegraphics[width=1\linewidth]{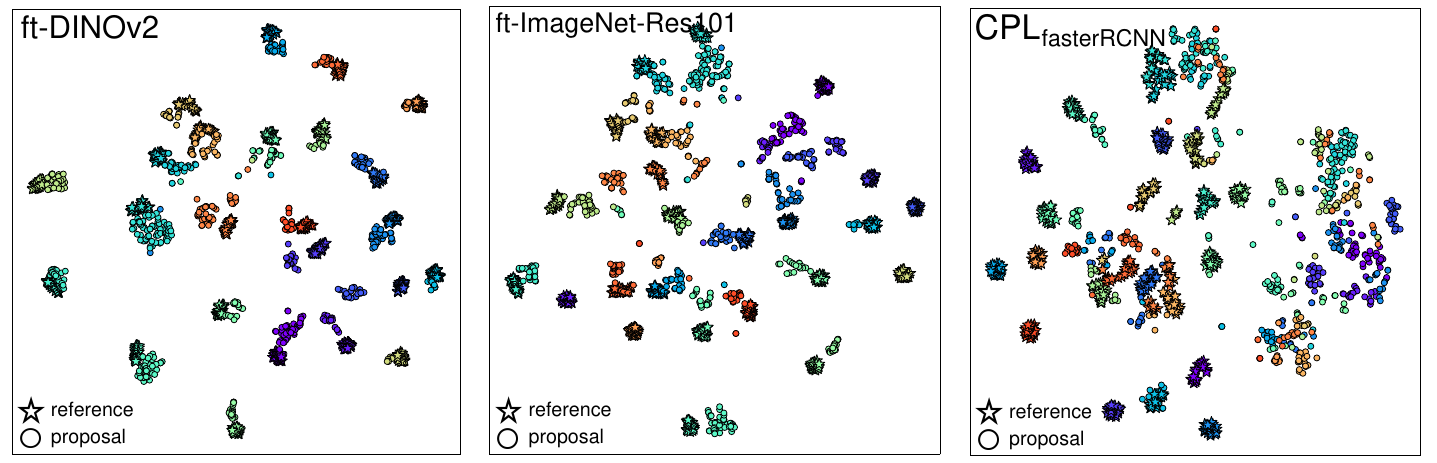}
\vspace{-7mm}
\caption{\small {\bf t-SNE visualization of features} learned by different methods.
We compare our finetuned DINOv2 against
a finetuned ImageNet-pretrained ResNet101 model and the baseline instance detector CPL with a FasterRCNN architecture.
We color feature points w.r.t instance IDs and overlay star-marked visual references (zoom-in to see better).
Visually,  the finetuned DINOv2 extracts more discriminative features.
}
\vspace{-2mm}
\label{fig:t-SNE}
\end{figure*}

{\bf Implementation details.} 
We carry out all experiments on a single NVIDIA 3090 GPU and implement methods with PyTorch.
Unless otherwise specified, our IDOW uses GroundingDINO~\cite{liu2023grounding} as the proposal detector and DINOv2~\cite{oquab2023dinov2} as the feature foundation model to finetune.
We use the pretrained [CLS] token to extract features and 
class embedding to compute similarity scores for matching.
When finetuning, we set hyper-parameter $\alpha=0.5$ in the metric learning loss. We use Adam optimizer with a learning rate 1e-3 and weight decay 0.5. 
We set the batch size as 100 and finetune the model for 10 epochs.

{
\setlength{\tabcolsep}{0.1em} 
\begin{table}[t]
\centering
\small
\caption{\small
{\bf Benchmarking results in the NID setting} on RoboTools. 
We compare our IDOW with state-of-the-arts and make three salient conclusions. 
First, IDOW significantly outperforms previous methods, e.g., IDOW$_{\text{GroundingDINO}}$ (59.4 AP) > OTS-FM$_{\text{GroundingDINO}}$ (56.7 AP) > VoxDet (18.7 AP).
This demonstrates the effectiveness of approaching InsDet from the open-world perspective.
Second, adapting FMs by our IDOW further boosts the performance by 3-5 AP, e.g., IDOW$_{\text{SAM}}$ (51.9 AP) > OTS-FM$_{\text{SAM}}$ (46.5 AP).
Third, adopting stronger FMs achieves better performance, e.g., IDOW$_{\text{GroundingDINO}}$ (59.4 AP) $>$ IDOW$_{\text{SAM}}$ (51.9 AP) in IDOW.
}
\vspace{-2mm}
\begin{tabular*}{\hsize}{@{}@{\extracolsep{\fill}}llccccc@{ }}
\toprule
{\bf Method} & \textcolor{grey}{\bf Venue \& Year} & {\bf AP} & {\bf AP$_{50}$} & {\bf AP$_{75}$} \\
\midrule
OS2D~\cite{osokin2020os2d} & \textcolor{grey}{ECCV 2020} 
& \cellcolor{col3}2.9 & 6.5 & 2.0
\\
DTOID~\cite{mercier2021deep} & \textcolor{grey}{WACV 2021}
& \cellcolor{col3}3.6 & 9.0 & 2.0
\\
OLN$_\text{Corr.}$~\cite{kim2022learning} & \textcolor{grey}{RA-L 2022}
& \cellcolor{col3}14.4 & 18.1 & 15.7
\\
VoxDet~\cite{li2024voxdet} & \textcolor{grey}{NeurIPS 2023}
& \cellcolor{col3}18.7 & 23.6 & 20.5 
\\
OTS-FM$_{\text{SAM}}$ \cite{shen2023high} & \textcolor{grey}{NeurIPS 2023}
& \cellcolor{col3}46.5 & 55.9 & 50.4
\\
OTS-FM$_{\text{GroundingDINO}}$ \cite{shen2023high} & \textcolor{grey}{NeurIPS 2023}
& \cellcolor{col3}56.7 & 64.8 & 59.0
\\
\midrule
IDOW$_{\text{SAM}}$ & & \cellcolor{col3}51.9 & 63.8 & 56.5
\\ 
IDOW$_{\text{GroundingDINO}}$ & 
& \cellcolor{col3}\bf{59.4} & \bf{67.8} & \bf{61.8} 
\\ 
\bottomrule
\end{tabular*}
\vspace{-2mm}
\label{tab:robotools-NID}
\end{table}
}

\subsection{Benchmarking Results}
{\bf Quantitative results.}
We evaluate our method against prior approaches on the InsDet and RoboTools dataset.
Table~\ref{tab:insdet-CID} and \ref{tab:robotools-NID} list detailed results in the CID and NID settings, respectively.
We summarize three salient observations:

First, {\it approaching InsDet from the open-world perspective significantly outperforms previous methods.}
From both CID and NID settings, 
we find that, despite not being explicitly trained on the instances of interests, OTS-FM already achieves $>$10 AP than traditional detector-based methods which particularly train on such instances.
Adapting FMs by our IDOW boosts performance by 3-5 AP.

Second, {\it IDOW significantly outperforms state-of-the-art InsDet methods}, achieving over 5 AP higher on average than methods like OTS-FM in both CID and NID settings, demonstrating the effectiveness of our techniques.

Third, {\it IDOW is versatile and applicable to different open-world pretrained FMs.}
By comparing IDOW$_{\text{SAM}}$ and IDOW$_{\text{GroundingDINO}}$, we find that adopting stronger FMs (GroundingDINO vs. SAM) achieves better performance.

{\bf Qualitative Results.}
Fig.~\ref{fig:closed-world-vis}  and \ref{fig:open-world_RoboTools} compares InsDet results by different methods in the CID and NID settings, respectively.
We can see that our IDOW  has higher precision and detects  more wanted instances, which are small and under low lighting conditions.
Fig.~\ref{fig:t-SNE} uses tSNE to visualize features of visual references and detected proposals (in the HR-InsDet dataset) computed by our finetuned DINOv2 (ft-DINOv2) in the left panel, and other non-FM methods
including a finetuned ImageNet-pretrained ResNet101 backbone (ft-ImageNet-Res101) and CPL$_{\text{FasterRCNN}}$ in the middle and right panels, respectively. 
Clearly, ft-DINOv2 matches proposals and references much better than other models.

\subsection{Ablation Study and Further Analysis}
\begin{figure}[t]
\centering
\includegraphics[width=0.95\linewidth]{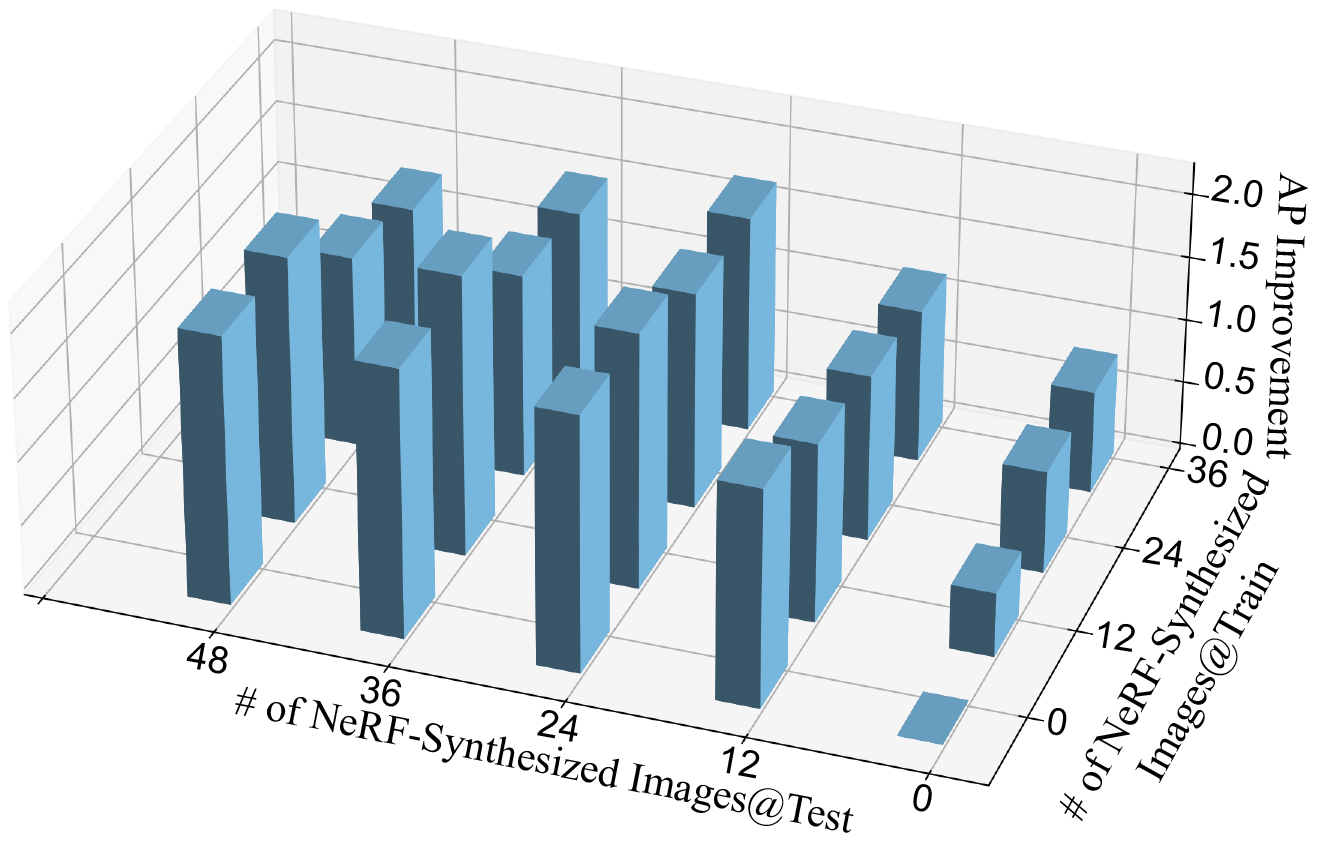}
\vspace{-2mm}
    \caption{\small 
    {\bf 
    Analysis of the number of added NeRF-synthesized images in training and testing w.r.t AP improvement} over the ``Train'' method (Table~\ref{tab:ablation}), which by default finetunes DINOv2 using the available visual references (24 per instance).
    We carry out this analysis in the CID setting on the HR-InsDet dataset.  
    On the one hand, adding NeRF-synthesized images helps finetune FM for InsDet, but adding more (e.g., $>$24) hurts the performance, likely due to overfitting to the artifacts of NeRF-synthesized images.
    On the other hand, adding more NeRF images in testing consistently improves performance.
    }
    \label{fig:ablation-num_views}
\vspace{-2mm}
\end{figure}

In Table~\ref{tab:ablation}, we ablate the proposed techniques of NeRF-synthesis and distractor sampling.
Results shows that adapting the FM DINOv2 greatly boosts InsDet performance (from 51.68 AP by the baseline to 53.94 AP).
Importantly, including NeRF-synthesized images in both training and testing significantly improves the performance.
Yet, using them in testing brings more performance gains than training (refer to the next study in Fig.~\ref{fig:ablation-num_views}).
Eventually, applying our techniques boosts AP to 57.01.
The supplemental material contains more quantitative and qualitative results.

{
\setlength{\tabcolsep}{0.1em} 
\begin{table*}[t]
\centering
\small
\caption{\small
{\bf Ablation study of each strategy involved in our IDOW.}
We carry out the study in the CID setting on the HR-InsDet dataset.
We use OTS-FM$_{\text{GroundingDINO}}$ as a baseline, over which we incrementally add each strategy.
``Train'' means foundation model adaptation through finetuning on the available data.
{\bf DA} denotes {\bf D}ata {\bf A}ugmentation with NeRF-generated novel-views;
{\bf DS} denotes {\bf D}istractor {\bf S}ampling.
Results clearly demonstrate that all the four strategies help achieve better InsDet performance.
Finetuning FMs on the given visual references enhances detection performance, cf. Train (53.94 AP) $>$ baseline (51.68 AP).
Moreover, NeRF-based data augmentation improves the final detection performance, particularly when used in testing, cf. Train+$\text{DA@Test}$ (56.44 AP) $>$ Train+$\text{DA@Train}$ (54.48 AP) $>$ baseline (51.68 AP).
Lastly, applying distractor sampling (DS) improves the final performance further, cf. Train+DS (54.10 AP) $>$ Train (53.94 AP).
}
\vspace{-3mm}
\begin{tabular*}{\hsize}{@{}@{\extracolsep{\fill}}cccccccccccc@{ }}
\toprule
\multicolumn{4}{c}{\bf Strategies} & \multicolumn{6}{c}{\bf AP} & {\bf AP$_{50}$}  & {\bf AP$_{75}$} \\
\cmidrule(l){1-4} \cmidrule(l){5-10} 
$\text{DA@Test}$ & Train & $\text{DA@Train}$ & DS & {\tt avg} & {\tt hard} & {\tt easy} & {\tt small} & {\tt medium} & {\tt large} & & \\
\midrule
\multicolumn{4}{c}{baseline: OTS-FM$_{\text{GroundingDINO}}$}
& \cellcolor{col3}51.68 & 37.23 & 58.72
& 28.79 & 58.55 & 69.22
& 62.50 & 56.78
\\ 
\midrule
& \checkmark & &
& \cellcolor{col3}53.94 & 37.54 & 61.52
& 30.18 & 60.98 & 71.60
& 65.18 & 59.36
\\ 
& \checkmark & \checkmark &
& \cellcolor{col3}54.48 & 38.31 & 61.71
& 31.18 & 61.10 & 72.55
& 66.06 & 59.98
\\ 
& \checkmark & & \checkmark 
& \cellcolor{col3}54.10 & 37.73 & 61.65
& 30.27 & 61.19 & 71.27
& 65.37 & 59.57
\\ 
& \checkmark & \checkmark & \checkmark
& \cellcolor{col3}54.92 & 38.28 & 62.41
& 32.00 & 61.61 & 74.21 
& 66.58 & 60.50
\\ 
\midrule
\checkmark & & &
& \cellcolor{col3}55.24 & 39.64 & 62.81
& 34.00 & 61.20 & 72.38
& 66.99 & 60.88
\\ 
\checkmark & \checkmark & &
& \cellcolor{col3}56.44 & 39.71 & 64.24
& 34.72 & 63.13 & 74.42
& 68.50 & 62.31
\\ 
\checkmark & \checkmark & \checkmark &
& \cellcolor{col3}56.92 & 39.85 & 64.85
& 34.66 & 62.74 & 75.48
& 68.89 & 62.80
\\ 
\checkmark & \checkmark & &\checkmark 
& \cellcolor{col3}56.51 & 39.70 & 64.36
& 34.34 & 61.65 & 75.89
& 68.71 & 62.38
\\ 
\checkmark & \checkmark & \checkmark & \checkmark
& \cellcolor{col3}57.01 & 40.74 & 64.36
& 35.25 & 62.98 & 73.64 
& 69.33 & 62.84 
\\
\bottomrule
\end{tabular*}
\vspace{-2mm}
\label{tab:ablation}
\end{table*}
}

Following the above, 
we study the affects of using different number of NeRF-synthesized images in training and test.
Fig.~\ref{fig:ablation-num_views} depicts the results.
(1) {\it Adding NeRF-generated images helps foundation model adaptation for InsDet}. 
Specifically, adding  $<$24 generated images boosts the performance, especially when 0 synthesized images are used in testing.
However, adding $>$24 generated images for training induces diminishing performance gains, likely because that the adapted FM overfits to artifacts in NeRF generated images. 
(2) {\it Adopting more synthetic images in testing consistently improves performance.}
Yet, the performance by adding more (e.g., $>$36) in testing saturates.
The peak performance occurs when 36 NeRF-synthesized images are used in testing.

\section{Discussions}
\label{sec:discussion}

{\bf Broader Impacts.}
Approaching InsDet in the open-world has a significant impact on various downstream applications, such as robotics and AR/VR. 
Detecting object instances in RGB images is considered the first step of contemporary perception algorithms. A stronger InsDet model provides richer information for the following steps, such as grasping or trajectory forecasting in robotics. 
Additionally, our work shows the promising performance of adapting FMs to InsDet, which will foster future research in studying different properties of FMs, such as adapting FMs to other tasks.
Nonetheless, our research utilizes FMs that are potentially pretrained on private data. At this point, it is unclear how such FMs might have bias or ethical issues.
As a result, we have no safeguard measure to ensure that the finetuned/adapted FMs are fair, even though we find no visible issues in the image-only dataset used for finetuning.
Further, adopting SAM may result in missed detections of critical instances relevant to specific downstream applications, such as medical devices or other rare objects. Such potential issues are not addressed in this work.

{\bf Limitations and future work.}
Experiments demonstrate that our proposed techniques effectively leverage FMs to address InsDet across various settings. 
Nonetheless, we note several limitations for future improvements.
First, the current pipeline relies on the proposals generated from GroundingDINO. Despite its high recall, it is not yet ready for real-time applications.
Future work should consider faster proposal detection methods.
Second, the NeRF augmentation step is also slow since it requires training a NeRF for every instance, which takes 1 hour and 500MB space using one NVIDIA 3090 GPU. While developing efficient NeRF methods is beyond our scope, we point out that efficient NeRF methods can be readily used in our pipeline, such as~\cite{muller2022instant} which trains a NeRF in seconds and~\cite{chen2023single} which can train a single NeRF for multiple objects.

\section{Conclusion}

Instance detection is highly applicable in various scenarios such as in robotics and AR/VR.
We elaborate on its open-world nature, necessitating the development of InsDet models in the open world.
We solve InsDet from the open world perspective, embracing foundation models (FMs) and data sampled in the open world.
In particular, we propose simple and effective techniques such as metric learning, distractor sampling, and novel-view synthesis.
Extensive experiments in different settings validate that our techniques in adapting FMs significantly boost InsDet performance.

{
\small
\bibliographystyle{ieeenat_fullname}
\bibliography{main}
}

\clearpage
\maketitlesupplementary

\renewcommand{\thesection}{\Alph{section}}
\setcounter{section}{0}

\section*{}
\begin{center}
    \emph{\bf \em \large Outline}
\end{center}
In this supplemental material, we provide additional experimental results,
more detection visualizations, and open-source code. 
Below is the outline of this document.

\begin{itemize} [noitemsep, topsep=-1pt, leftmargin=*]    
\item
    {\bf Section~\ref{sec:jupyter-notebook}}.
    We provide demo code for our proposed method IDOW in both CID and NID using Jupyter Notebook.
\item 
    {\bf Section~\ref{sec:ablation}}. We conduct additional ablation studies of each strategy involved in our IDOW.
\item 
    {\bf Section~\ref{sec:visual}}. We provide additional qualitative visualizations of detection results from different methods.
\end{itemize}

\section{Open-Source Code}
\label{sec:jupyter-notebook}
In the project page (\url{https://shenqq377.github.io/IDOW/}),
we release open-source code in the form of Jupyter Notebook plus Python files.

{\bf Why Jupyter Notebook?}
We prefer to release the code using Jupyter Notebook (\url{https://jupyter.org}) because it allows for interactive demonstration for education purposes.
In case the reader would like to run Python script, using the following command can convert a Jupyter Notebook file {\tt xxx.ipynb} into a Python script file {\tt xxx.py}:
{\small 
\begin{verbatim}jupyter nbconvert --to script xxx.ipynb\end{verbatim}
}

{\bf Requirement}.
Running our code requires some common packages.
We installed Python and most packages through Anaconda. A few other packages might not be installed automatically, such as Pandas, torchvision, and PyTorch, which are required to run our code. Below are the versions of Python and PyTorch used in our work. 
\begin{itemize}
\item Python version: 3.9.16 [GCC 7.5.0]
\item PyTorch version: 2.0.0
\end{itemize}
We suggest assigning $>$30GB space to run all the files.

{\bf License}.
We release open-source code under the MIT License to foster future research in this field.

{\bf Demo.}
The Jupyter Notebook files below demonstrate our proposed method IDOW in both CID and NID settings. During the training stage, we finetune DINOv2 with visual references from HR-InsDet/RoboTools dataset (in the CID setting) or OWID dataset (in the NID setting). We use GroundingDINO to detect instance-agnostic proposals for a given testing image. We feed these proposals into DINOv2 for feature representation, just like how we represent visual references of object instances. Over the features, we use stable matching on the cosine similarities between proposals and visual references to find the best match, yielding the final detection results.

\begin{itemize}

\item
\begin{verbatim}demo_CID_InsDet.ipynb\end{verbatim}
Running this file finetune DINOv2 with visual references of 100 object instances from HR-InsDet dataset, and compute feature representation for GroundingDINO-detected proposals and visual references. The final detection is still performed using the stable matching algorithm with cosine similarities. This file presents the proposed approach IDOW$_{\text{GroundingDINO}}$ in the CID setting.

\item
\begin{verbatim}demo_NID_RoboTools.ipynb\end{verbatim}
Running this file finetune DINOv2 with visual references of 9691 object instances from OWID dataset, and compute feature representation for GroundingDINO-detected proposals and visual references. The final detection is still performed using the stable matching algorithm with cosine similarities. This file presents the proposed approach IDOW$_{\text{GroundingDINO}}$ in the NID setting.
\end{itemize}

\section{Ablation Study and Further Analysis}
\label{sec:ablation}
We include additional ablation studies to supplement the results in the main paper. Specifically, we study: (1) performance of IDOW w.r.t different loss functions; (2) correlation plot between open-world sampled distractors and visual references; (3) performance of IDOW w.r.t different model backbones. Lastly, we present quantitative results evaluated by average recall (AR) and precision-recall curves of different methods to better understand the performance improvements of our proposed approach.

{
\setlength{\tabcolsep}{0.75em} 
\begin{table*}[t]
\centering
\small
\caption{\small
{\bf Comparisons with different losses in foundation model adaptation.} 
We carry out the study on the HR-InsDet dataset in both CID and NID settings. Clearly, adapting FMs with triplet loss performs better than finetuning the model either by a contrastive loss or a cross entropy (CE) loss. This could be attributed to the advantages of triplet loss in handling hard examples in IDOW. }
\vspace{-2mm}
\begin{tabular}{llcccccccc}
\toprule
{\bf Setting} & {\bf Loss} & \multicolumn{6}{c}{\bf AP} & {\bf AP$_{50}$}  & {\bf AP$_{75}$} \\
\cmidrule(l){3-8} 
 & & {\tt avg} & {\tt hard} & {\tt easy} & {\tt small} & {\tt medium} & {\tt large} & & \\
\midrule 
\multirow{3}{*}{CID} 
& CE Loss
& \cellcolor{col3}55.72 & 39.12 & 63.02
& 34.16 & 60.66 & 72.23
& 67.70 & 61.41
\\
& Contrastive Loss
& \cellcolor{col3}53.94 & 38.65 & 60.89
& 32.11 & 58.86 & 71.07
& 65.46 & 59.34
\\
& Triplet Loss
& \cellcolor{col3}57.01 & 40.74 & 64.36
& 35.25 & 62.98 & 73.64 
& 69.33 & 62.84
\\
\midrule
\multirow{3}{*}{NID} 
& CE Loss
& \cellcolor{col3}55.52 & 39.94 & 62.58
& 34.09 & 61.41 & 72.83 
& 67.30 & 61.19 
\\
& Contrastive Loss
& \cellcolor{col3}55.23 & 39.64 & 62.38
& 33.75 & 61.38 & 71.93
& 67.19 & 60.91
\\
& Triplet Loss
& \cellcolor{col3}56.01 & 40.42 & 63.36
& 35.14 & 62.22 & 72.55 
& 68.11 & 61.75
\\
\bottomrule
\end{tabular}
\label{tab:Loss}
\end{table*}
}

\begin{figure*}[t]
\centering
\includegraphics[width=0.99\textwidth]{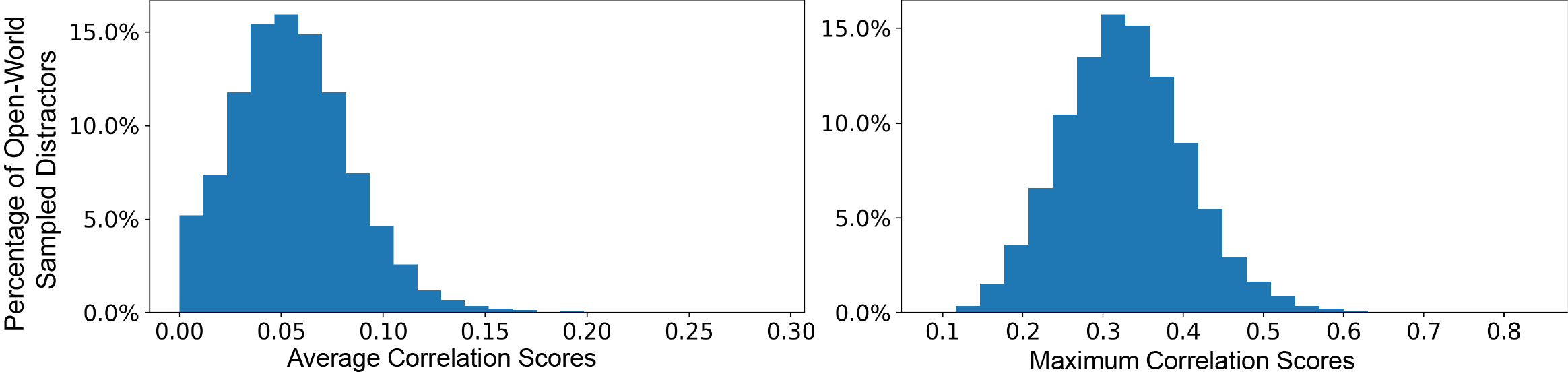}
\vspace{-3mm}
\caption{\small 
{\bf Correlation plot between open-world sampled distractors and visual references.}
We plot the correlation measured by average (left) and maximum (right) cosine similarity between sampled distractors and visual references on the HR-InsDet dataset. 
The average score measures the average distance between each distractor and all visual references while the maximum score finds the highest score.
We find the majority of distractors have high enough maximum correction scores ($\geq$ 0.3), indicating open-world sampled distractors are “hard” for FMs to distinguish from at least one instance visual reference. 
This observation, together with our hard example sampling strategy, demonstrates the effectiveness of approach InsDet with open-world sampled distractors.
}
\label{fig:correlation_bg-profile}
\end{figure*}

{
\setlength{\tabcolsep}{0.1em} 
\begin{table*}[t]
\centering
\small
\caption{\small
{\bf Ablation study of different DINOv2 backbone involved in our IDOW}. We carry out the study in the CID setting on the HR-InsDet dataset. We compare different input resolutions and DINOv2 backbones, and make two salient conclusions. First, stronger backbones lead to better performance. Second, the increasing input resolution boosts the detection performance.
}
\vspace{-2mm}
\begin{tabular*}{\hsize}{@{}@{\extracolsep{\fill}}lccccccccccc@{ }}
\toprule
{\bf DINOv2 backbone} & {\bf \# Param} & {\bf Input size} & \multicolumn{6}{c}{\bf AP} & {\bf AP$_{50}$}  & {\bf AP$_{75}$}\\  
\cmidrule(l){4-9} 
 & & & {\tt avg} & {\tt hard} & {\tt easy} & {\tt small} & {\tt medium} & {\tt large} & &\\ 
\midrule 
ViT-s/14 & 21M & $224\times224$
& \cellcolor{col3}54.00 & 36.49 & 62.48
& 33.10 & 61.18 & 72.80
& 65.18 & 59.84 
\\
ViT-s/14 & 21M & $448\times448$
& \cellcolor{col3}56.84 & 40.67 & 64.49
& 35.22 & 62.54 & 74.01
& 68.85 & 62.90 
\\
ViT-s/14 & 21M & $518\times518$
& \cellcolor{col3}57.01 & 40.74 & 64.36
& 35.25 & 62.98 & 73.64 
& 69.33 & 62.84 
\\
ViT-b/14 & 86M & $518\times518$
& \cellcolor{col3}58.63 & 41.45 & 66.41
& 36.92 & 65.11 & 77.07
& 71.28 & 64.78 
\\
\bottomrule
\end{tabular*}
\label{tab:ablation-DINOv2Backbone}
\end{table*}
}

{
\setlength{\tabcolsep}{0.55em} 
\begin{table*}[t]
\centering
\small
\caption{\small{\bf Benchmarking results w.r.t average recall (AR) for \emph{small}, \emph{medium} and \emph{large} instances}.
``AR@max10'' means AR within the top-10 ranked detections. 
In computing AR, we rank detections by using the detection confidence scores of the learning-based methods (e.g., FasterRCNN) or similarity scores in the non-learned methods OTS-FM.
AR$_s$, AR$_m$, and AR$_l$ are breakdowns of AR for small, medium, and large testing object instances. 
Results show that (1) SAM or GroundingDINO generally recalls more instances than others, (2) methods with high AR typically achieve better AP (cf. Table~\ref{tab:insdet-CID}), and (3) all methods suffer from small instances. 
}
\vspace{-2mm}
\begin{tabular}{lccccc}
\toprule
& {\bf AR@max10} & {\bf AR@max100} & {\bf AR$_s$@max100} & {\bf AR$_m$@max100} & {\bf AR$_l$@max100} \\
\midrule
CPL$_{\text{FasterRCNN}}$~\cite{NIPS2015_fasterRCNN, dwibedi2017cut}
& 26.24 & 39.24
& 14.83 & 44.87 & 60.05
\\
CPL$_{\text{RetinaNet}}$~\cite{lin2017focal, dwibedi2017cut}
& 26.33 & 49.38
& 22.04 & 56.76 & 69.69
\\
CPL$_{\text{CenterNet}}$~\cite{dwibedi2017cut, zhou2019objects}
& 23.55 & 44.72
& 17.84 & 52.03 & 64.58
\\
CPL$_{\text{FCOS}}$~\cite{dwibedi2017cut, tian2019fcos}
& 25.82 & 46.28
& 22.09 & 52.85 & 64.11
\\
CPL$_{\text{DINO}}$~\cite{dwibedi2017cut, zhang2022dino}
& 29.84 & 54.22
& 32.00 & 59.43 & 72.92
\\
OTS-FM$_{\text{SAM}}$~\cite{shen2023high, kirillov2023segment}
& 40.02 & 63.06
& 31.11 & 70.40 & 90.36
\\
\midrule
IDOW$_{\text{GroundingDINO}}$
& 40.29 & 77.09
& 53.53 & 83.73 & 94.06\\
\bottomrule
\end{tabular}
\vspace{0mm}
\label{tab:RealDB-AR}
\end{table*}
}

\begin{figure*}[!tbp]
\centering
\small
\ \hspace{0mm} (a) CID, HR-InsDet \hspace{55mm} (b) NID, RoboTools 
\\
\includegraphics[width=0.49\textwidth, trim={0 0 0 0},clip]{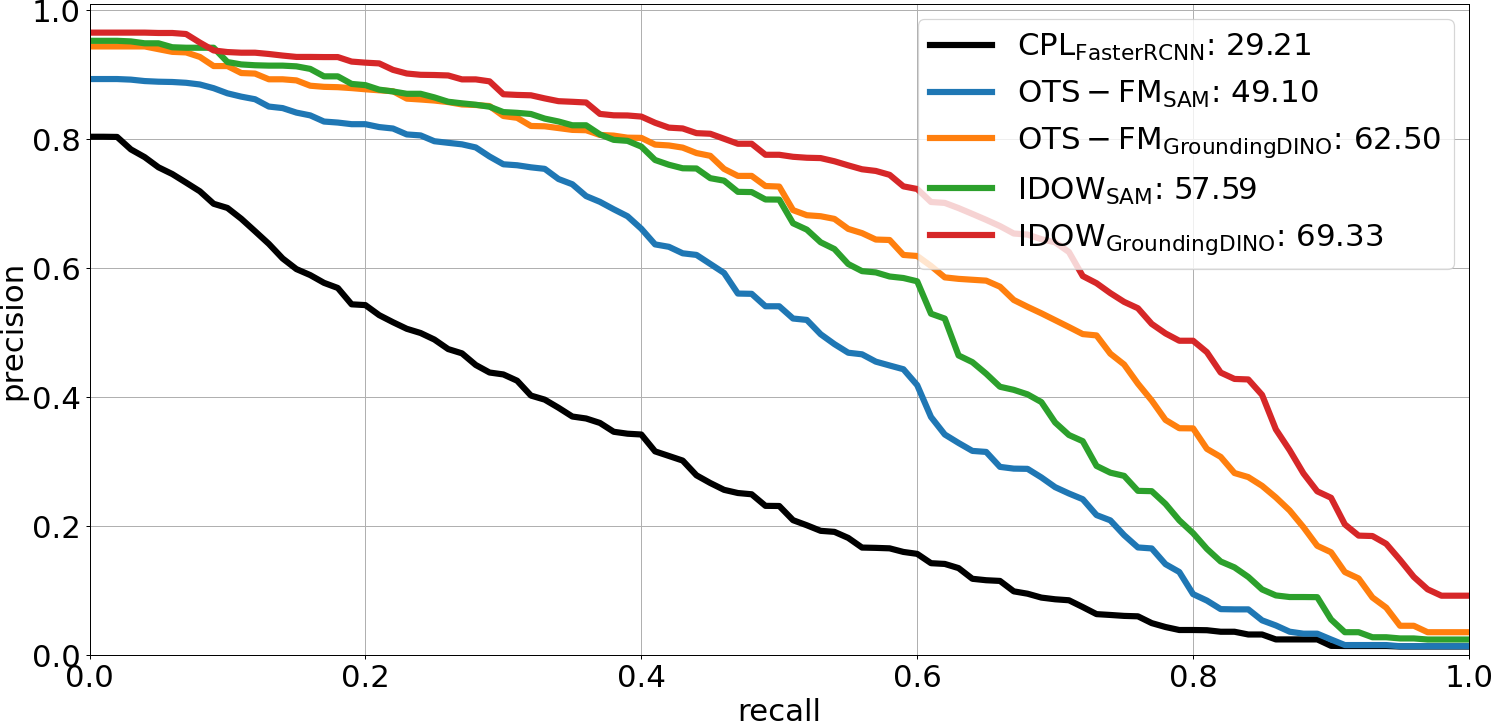}
\includegraphics[width=0.49\textwidth, trim={0 0 0 0},clip]{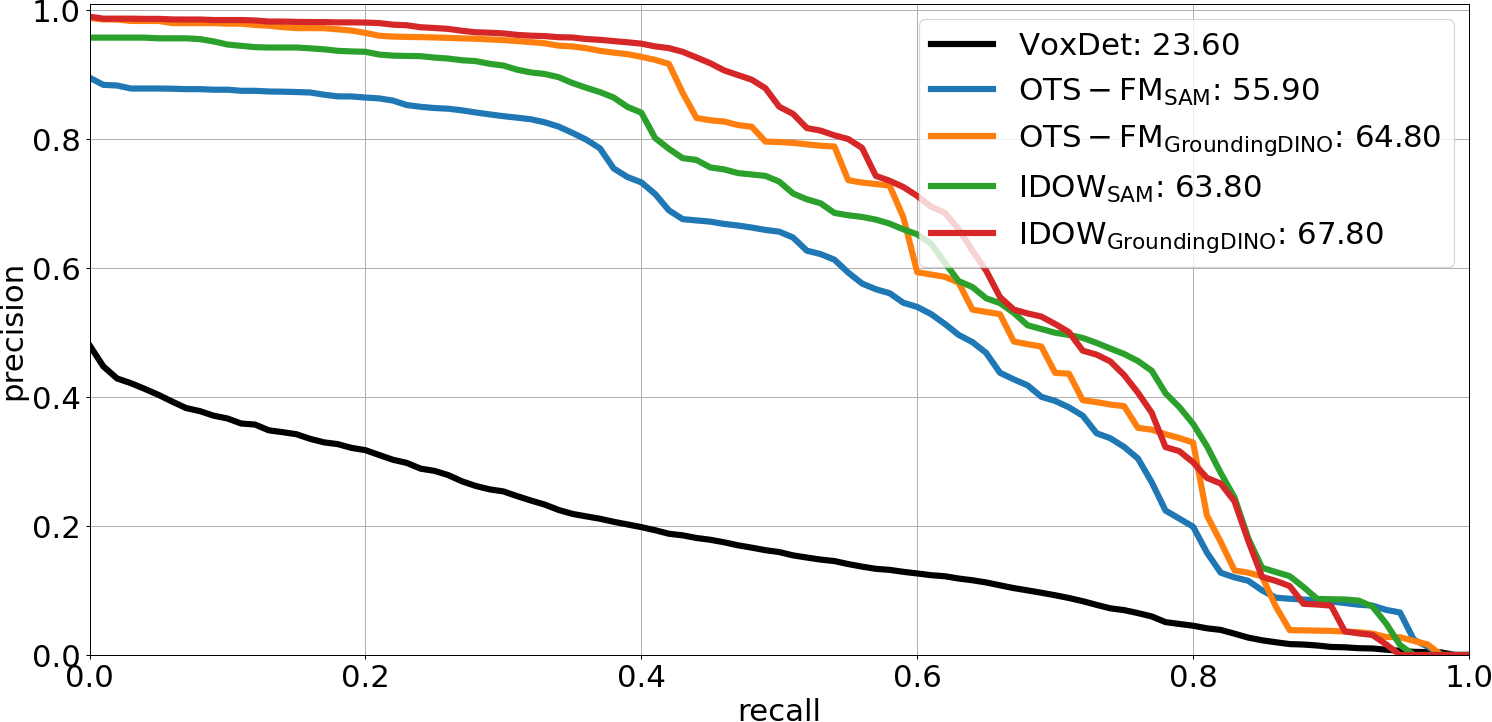}
\vspace{-2mm}
\caption{\small 
{\bf Precision-recall curves with IoU=0.5
(AP$_{50}$) under both CID and NID settings.}
We find that (1) our proposed method IDOW outperforms the previous state-of-the-art OTS-FM under both settings in terms of both recall and precision, and (2) IDOW maintains more stable precision accuracy when increasing recall (e.g., IDOW$_{\text{GroundingDINO}}$ v.s. OTS-FM$_{\text{GroundingDINO}}$), indicating that it has higher stability and stronger robustness in distinguishing between positive and negative samples.
}
\label{fig:pr_curve}
\end{figure*}

{\bf Performance of IDOW w.r.t different loss functions.}
There are various loss functions in deep learning to learn effective feature representations, each designed with a particular goal, e.g., learn features that discriminate between classes by cross entropy (CE) loss~\cite{pouyanfar2018survey}, structure the feature space based on similarity and dissimilarity by contrastive loss~\cite{chen2020simple,tabassum2022hard}, etc. 
We ablate three choices of finetuning FMs in IDOW, specifically CE loss, contrastive loss and triplet loss, in this supplement. 
To keep the comparison fair, we keep all the settings same except the loss function. 
As shown in Table~\ref{tab:Loss}, we find that adapting FMs with triplet loss performs better than finetuning the model with either a contrastive loss or a cross-entropy loss in both CID and NID settings, e.g., 55.52 AP by CE loss vs. 56.01 AP by triplet loss in NID. This could be attributed to the advantages of triplet loss in handling hard examples in IDOW. 

{\bf Correlation plot between open-world sampled distractors and visual references.}
The experimental results in the main paper show that leveraging open-world sampled distrators leads to performance improvements.
However, it is less clear how these distractors correlate with visual references and thus contribute to better adapt FMs for InsDet.
We plot the correlations measured by cosine similarities between distractors and visual references. Specifically, we report average and maximum cosine similarity scores. The former one describes the average score of each distractor w.r.t all visual references while the later one represents the maximum score. 
As shown in Fig.~\ref{fig:correlation_bg-profile}, the majority of distractors have maximum correction scores $\geq$ 0.3 despite the average score is low. 
This observation suggests that the majority of distractors are ``hard'' for FMs to distinguish from at least one instance. This explains adding open-world sampled distractors improve the performance together with our hard example sampling strategy.
Nonetheless, it is worth for future work to explore how to better extract and leverage these open-world distractors.

{\bf Choice of FM backbones and input resolutions in IDOW.}
The experiments present in the main paper demonstrate the effectiveness of IDOW using DINOv2 with ViT small (ViT-s) as the backbone.
Here, we explore the performance change of IDOW by adopting different DINOv2 backbones. According to \cite{oquab2023dinov2}, authors increase the resolution of images to $518\times518$ during a short period at the end of pretraining.
Additionally, input resolutions also correlates with the performance of FMs from the recent literature~\cite{shen2023high}. Therefore, we further explore the effect of increasing input resolutions in Table~\ref{tab:ablation-DINOv2Backbone}. 
We summarize two conclusions: 
{\it (1) Stronger backbones lead to better performance.} Comparing ViT-s and ViT-b, it is clear that adopting a deeper backbone achieves $>$ 4 AP improvements. 
{\it (2) Input resolution matters and increasing resolution boosts the performance.} 
Comparing the ViT-s backbone with 224, 448 and 512 input resolutions, inputs with 512 $\times$ 512 shows $\sim$3 AP improvements over 224 $\times$ 224.

{\bf Quantitative results w.r.t average recall (AR).}
In addition to the experimental results of average precision (AP) in the main paper, we also report the average recall (AR) of different competitors as a supplement in Table~\ref{tab:RealDB-AR}. We carry out the experiments in the CID setting on HR-InsDet dataset. From Table~\ref{tab:RealDB-AR}, we make three conclusions. (1) Non-learned FMs are better at proposing instances, e.g., 40.29 AR in GroundingDINO vs. 26.24 AR in FasterRCNN. (2) Comparing with AP in Table~\ref{tab:insdet-CID}, methods with high AR typically achieve better AP. (3) All methods suffer from small instances.

{\bf Precision-recall curve of different methods.} 
We present the precision-recall curves of different competitors in the CID setting on the HR-InsDet dataset (ref. Table~\ref{tab:insdet-CID}) and in the NID setting on the RoboTools dataset (ref. Table~\ref{tab:robotools-NID}) in Fig.~\ref{fig:pr_curve}. 
We make two observations. (1) Our proposed method IDOW$_{\text{GroundingDINO}}$ outperforms the previous state-of-the-arts in terms of both recall and precision. 
(2) IDOW maintains more stable precision accuracy when increasing recall (e.g., IDOW$_{\text{GroundingDINO}}$ v.s. OTS-FM$_{\text{GroundingDINO}}$), indicating that it has higher stability and stronger robustness in distinguishing between positive and negative samples.

{\bf Runtime/Inference speed.} 
We provide inference time of our IDOW$_{\text{GroundingDINO}}$ averaged over HR-InsDet / RoboTools testing images w.r.t each step in Table~\ref{tab:inference_speed}: proposal detection using GroundingDINO, feature extraction with DINOv2, and feature matching including proposal-reference similarity computation and stable matching. In HR-InsDet / RoboTools, testing image resolution is 1024x2048 / 1080x1920; each instance has 24 / 100 visual references, with GroundingDINO generating 40 / 35 proposals per image on average. Our method is quite efficient in inference. Moreover, our hard example mining uses a simple min operation to focus on the hardest example in each training batch for each anchor. This min operation does not add additional compute cost.

\begin{figure*}[!t]
\centering
\small
\hspace{-5mm} Support \hspace{15mm} Ground Truth \hspace{18mm} Cut-Paste-Learn~\cite{dwibedi2017cut} \hspace{15mm} OTS-FM~\cite{shen2023high} \hspace{18mm} {\bf IDOW (ours)} \hspace{15mm} 
\\
\includegraphics[width=0.99\textwidth, trim={0 0 0 0},clip]{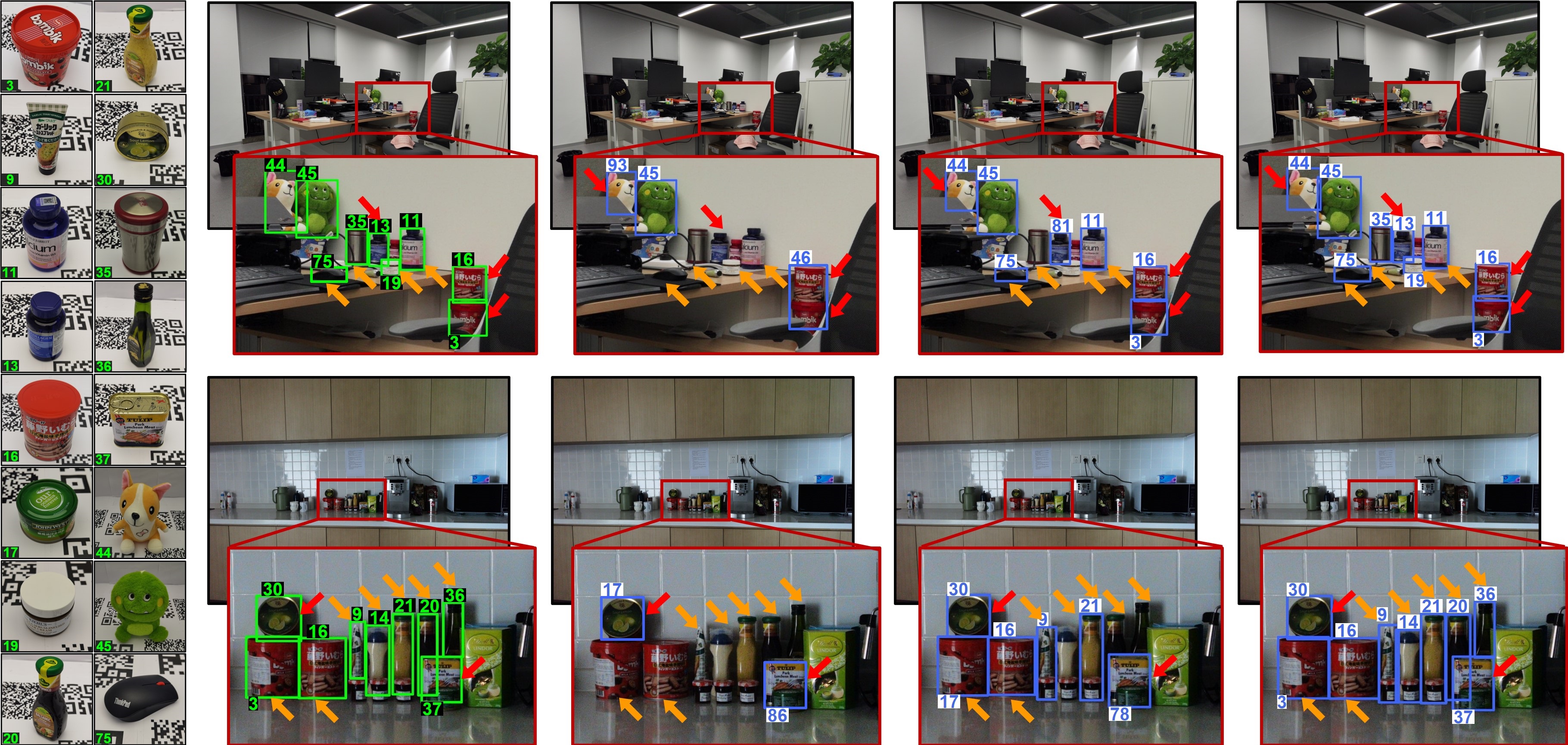}
\vspace{-2mm}
\caption{\small
{\bf Visual comparison of InsDet results by different methods in the CID setting} on HR-InsDet. We mark the ground-truth and predictions using \textcolor{darkgreen}{green} and {\setlength{\fboxsep}{1pt}\colorbox{lightblue}{blue}} boxes, respectively. Compared with Cut-Paste-Learn and OTS-FM, our IDOW detects more instances (see \textcolor{orange}{orange arrows}) with better accuracy (see \textcolor{purered}{red arrows}), when instances are partially occluded and illumination conditions change. Compared with OTS-FM, IDOW adapting a foundation model (i.e., DINOv2 used by
both) yields better features, enabling robust predictions in InsDet.}
\label{fig:CID_InsDet}
\vspace{-1mm}
\end{figure*}

\begin{figure*}[!t]
\centering
\small
\hspace{-5mm} Support \hspace{15mm} Ground Truth \hspace{21mm} VoxDet~\cite{li2024voxdet} \hspace{19mm} OTS-FM~\cite{shen2023high} \hspace{18mm} {\bf IDOW (ours)} \hspace{18mm} 
\\
\includegraphics[width=0.99\textwidth, trim={0 0 0 0},clip]{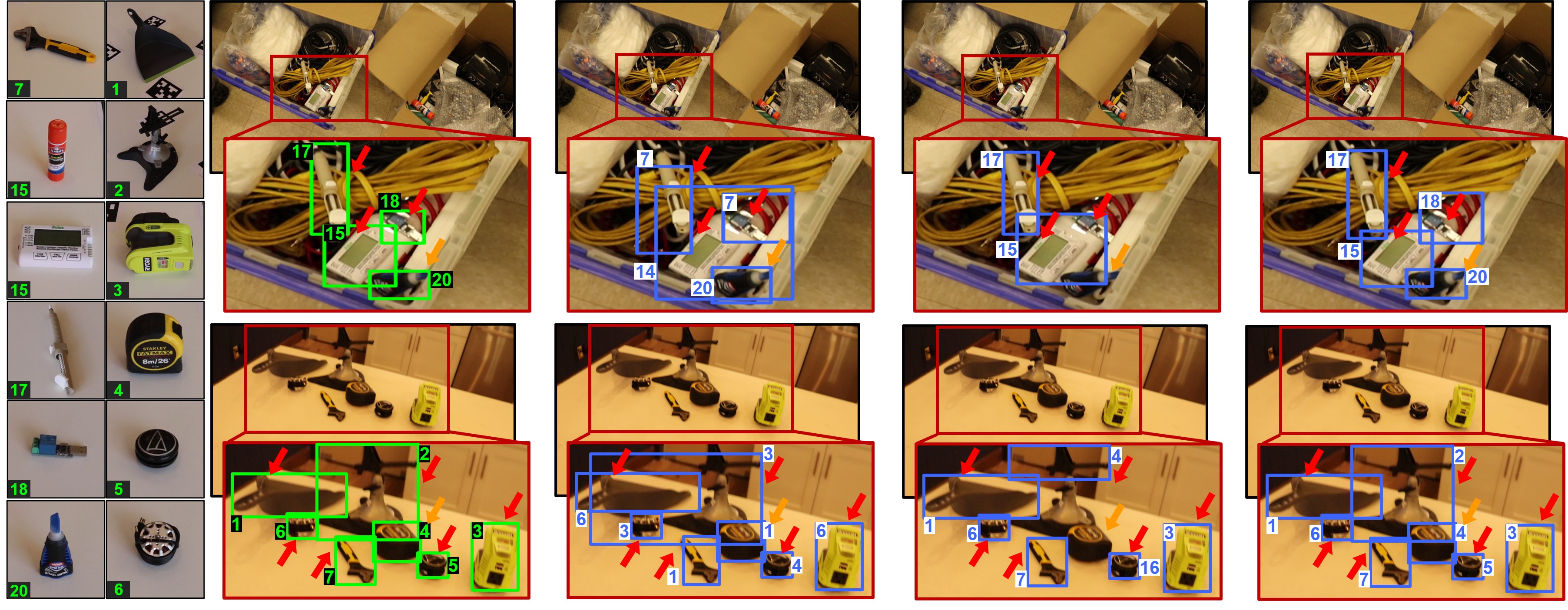}
\vspace{-2mm}
\caption{\small
{\bf Visual comparison of InsDet results by different methods in the NID setting} on RoboTools. We mark the ground-truth and predictions using \textcolor{darkgreen}{green} and {\setlength{\fboxsep}{1pt}\colorbox{lightblue}{blue}} boxes, respectively. Compared with VoxDet, our IDOW accurately detect instances (see \textcolor{purered}{red arrows}) in a blurred and cluttered testing scene. Compared with OTS-FM, our IDOW detects more instances (see \textcolor{orange}{orange arrows}).}
\label{fig:NID_InsDet}
\vspace{-1mm}
\end{figure*}

\section{Additional Visualizations}
\label{sec:visual}

{\bf Prediction visualizations.}
We present more detection comparisons under the CID setting on the HR-InsDet dataset in Fig.~\ref{fig:CID_InsDet}, and under the NID setting on the RoboTools dataset in Fig.~\ref{fig:NID_InsDet}. We attach instance IDs to ground-truth and predictions to highlight whether the instance recognition is correct compared to the visual references.
Our proposed method IDOW is compared with two previous arts, Cut-Paste-Learn~\cite{dwibedi2017cut} and OTS-FM~\cite{shen2023high}.
We observe that our proposed method is more robust under small size, similar appearance, pose variation and serve occlusion by embracing foundation models with NeRF augmentation. For example, in Fig.~\ref{fig:CID_InsDet}, although instances (No.3, No.13, No.44 in the top row, and No.9, No.14, No.20, No.21, No.36, No.37 in the bottom row) in HR-InsDet are partially distracted, our IDOW can accurately identify them. In the RoboTools benchmark, most instances are very small and placed with arbitrary pose variations in the cluttered scenes. Moreover, the testing scene images are blurred. From Fig.~\ref{fig:NID_InsDet}, we observe that our IDOW can accurately detect more instances than VoxDet and OTS-FM although the visual references are not seen during adaptation in the NID setting.

{
\setlength{\tabcolsep}{0.5em}
\begin{table}[t]
\centering
\small
\caption{\small{\bf Runtime (measured in seconds) of IDOW averaged over HR-InsDet / RoboTools testing images}.}
\vspace{-3mm}
\scalebox{0.9}{
\begin{tabular}{lccccccc}
\hline
 Dataset & proposal det. & fea. extraction & fea. matching & Total \\ 
\hline 
\multirow{1}{*}{HR-InsDet} & 0.123 & 0.018  & 0.042& 0.183 \\
\hline
\multirow{1}{*}{RoboTools} & 0.125 & 0.019 & 0.018 & 0.162 \\
\hline
\end{tabular}
}
\label{tab:inference_speed}
\vspace{-3mm}
\end{table}
}

\end{document}